\documentclass[10pt,conference]{IEEEtran}
\usepackage{times}  
\usepackage{helvet}  
\usepackage{courier}  
\usepackage[hyphens]{url}  
\usepackage{graphicx} 

\usepackage{amsthm,array,booktabs,tabularx,caption}
\usepackage[cmintegrals]{newtxmath}

\usepackage{algorithm}
\usepackage{algorithmic}
\usepackage{subcaption}
\theoremstyle{definition}

\theoremstyle{remark}

\theoremstyle{definition}

\usepackage[colorinlistoftodos]{todonotes}

\title{Learning to Bid Long-Term: Multi-Agent Reinforcement Learning with Long-Term and Sparse Reward in Repeated Auction Games}

\author{\IEEEauthorblockN{Jing Tan}
\IEEEauthorblockA{\textit{Huawei Technology} \\
Munich, Germany \\
jingtan@huawei.com}
\and
\IEEEauthorblockN{Ramin Khalili}
\IEEEauthorblockA{\textit{Huawei Technology} \\
Munich, Germany \\
ramin.khalili@huawei.com}
\and
\IEEEauthorblockN{Holger Karl}
\IEEEauthorblockA{\textit{Hasso Plattner Institute} \\
Berlin, Germany \\
holger.karl@hpi.de}
}

\begin{document}

\maketitle

\begin{abstract}
We propose a multi-agent distributed reinforcement learning algorithm that balances between potentially conflicting short-term reward and sparse, delayed long-term reward, and learns with partial information in a dynamic environment. We compare different long-term rewards to incentivize the algorithm to maximize individual payoff and overall social welfare. We test the algorithm in two simulated auction games, and demonstrate that 1) our algorithm outperforms two benchmark algorithms in a direct competition, with cost to social welfare, and 2) our algorithm's aggressive competitive behavior can be guided with the long-term reward signal to maximize both individual payoff and overall social welfare.\footnote{This paper is accepted for AAAI 2022 workshop ``Reinforcement Learning for Games''.}
\end{abstract}

\section{Introduction}
\label{sec:intro}

Auction is a common resource allocation mechanism in e.g.\ networking \cite{xu2012resource}\cite{xu2012interference}, energy \cite{lucas2013renewable}, e-commerce \cite{huang2011design}, for its efficient price discovery in a dynamic market with partial information \cite{schindler2011pricing}\cite{einav2018auctions}. In many such applications using auction mechanism, agents are designed to represent the bidders and automatically bid in the auctions. The agents have private information of goals and valuations and behave autonomously to maximize their individual payoff. The agents have learning capabilities to continuously improve their behavioral strategy and increase payoff in a dynamic environment, where the payoff depends not only on the environment, but also on other bidder agents' strategies \cite{busoniu2008comprehensive}. Ultimatedly, these agents learn by repeating the auction game to gradually maximize the long-term reward over time. 

Reinforcement learning (RL) algorithms are often used in such cases, for their ability to learn with sparse environment feedback and balance between exploitation and exploration \cite{teng2013reinforcement}\cite{almasri2020dynamic}. However, challenges remain. Firstly, although RL algorithms are often used to learn sequential tasks, many of them are still relatively ``short-term'' algorithms: learning is based on a reward given immediately after action and state transition, and as the prediction horizon extends farther into the future, influence of the current action decreases exponentially; moreover, if the reward is sparse and delayed, the reward estimation often has a high variance due to lack of predictable future states, especially with a big state-action space, and variance in the value of next states \cite{mataric1994reward}\cite{shahriari2017generic}. In the cases where decisions have long-term effects that are only apparent after a variable delay and where short-term rewards conflict with long-term goals, such ``short-term'' algorithms that maximize based on immediate reward would lead to worse performance in the long run. Secondly, many RL algorithms are designed for single agents. With multiple agents, the agents need to learn in a dynamic and competitive environment with partial information \cite{almasri2020dynamic}. Such decentralized decision making may lead to conflicts between social welfare (i.e.\ total reward of all agents) and individual gains, whereas the dynamic nature of the environment requires tradeoff between optimality and convergence while keeping computation and communication complexity tractable \cite{feigenbaum2007distributed}. 

To address these challenges, we propose DRACO2, a multi-agent, long-term learning algorithm with credit assignment. We define a reward mechanism that decouples short-term dense and long-term sparse rewards and enables learning on different time scales; we use credit assignment to break down the long-term reward into a weight vector that is aligned with short-term rewards. To learn in a dynamic and competitive environment, we use distributed RL. Specifically, our core RL algorithm learns the best-response strategy updated in a fictitious self-play (FSP) method. Our learning agents have a state- and reward-predictive model to improve prediction accuracy of the future and the consequence of their actions. Moreover, we use the curiosity-learning concept \cite{pathak2017curiosity}, which has an adversarial setting to encourage the RL model to explore state-action space where the agent lacks predictive power. 

To demonstrate the performance of DRACO2, we simulate two repeated auction games with learning-capable agents as bidders, and one single passive agent as a broker. The game setup is suitable for analyzing our algorithm, for it 1) creates a dynamic and competitive environment with independent agents, each with private values and goal; 2) has a vast state-action space; 3) creates conflicting short-term and long-term objectives: in the short term, the bidder is incentivized to receive immediate payoff, but in the long term, winning a bid would reduce the money available for future bids and bind the agent's resources, thus incurring opportunity cost; and 4) provides choices of long-term reward as incentive to bidders.

Empirical results show that DRACO2 outperforms both the short-term algorithm and the vanilla curiosity learning algorithm in a direct competition, although at the cost of reduced social welfare (i.e.\ total payoff received by all agents). To maximize social welfare without compromising individual gains, we use a fairness index score as long-term reward to replace the original profit-seeking goal. As a result, all agents with DRACO2 receive maximum cumulated payoff.

Our contributions:

\begin{itemize}
\item DRACO2 is extremely aggressive and competitive in both simulated auction games, outperforming benchmark algorithms, showing its capability to learn with sparse, delayed, sporadic reward and partial information in a dynamic, adversarial environment. 
\item Despite its aggressive behavior, it is easy to influence DRACO2 by simply replacing the profit-seeking goal with a fairness goal, compromising neither individual gain nor privacy. 
\item We open source our code \cite{biddinggame}.
\end{itemize}

\section{Related Work}
\label{sec:related}

One of the biggest challenges of applying RL in the real world is to learn behavior towards long-term goals with delayed and sparse reward signal \cite{dulac2021challenges}. One common approach is to extract features from historical records, thus linking the delayed reward to behaviors in the past \cite{hester2013texplore}. 
Learning with such algorithms is inefficient since learning from past experiences can only happen when the delayed outcomes become available. To address the delay, \cite{mann2018learning} factorizes one state into an intermediate and a final state with independent transition probabilities and predicts each state at different intervals. Reference \cite{hung2019optimizing} describes a credit-assignment method that focuses on the most relevant memory records via content-based attention; the algorithm is capable of locating past memory to execute new tasks and generalizes very well. These approaches focus more on the delay in reward signal and less on sparsity. In our setup, the long-term reward is delayed, sparse and sporadic.

To address sparsity of rewards, many model-based methods add intrinsic, intermediate rewards between sparse extrinsic reward signals. Such methods often adopt a supervised learning algorithm to predict next states and use the difference between the predicted and target state-action pair values as intrinsic reward. Although they propagate prediction inaccuracy into the future, they learn faster. For example, \cite{hester2013texplore} separately trains many ``feature models'' to predict each feature of the next state as well as a ``reward model'' to predict reward. Between sparse extrinsic rewards, the algorithm samples estimated next state and reward from the models. The models are only updated when there is new input available. Their approach assumes that state features are independent and can be learned separately, and the accuracy of the reward model is still related to the sparsity of the reward signal. 
\cite{pathak18largescale} uses a long-short-term memory (LSTM) to extract features from past memory that are more relevant to the current task, thus improving the model's generalization properties. The algorithm also uses two independent models to predict next state and action, the prediction accuracy becomes intermediate, intrinsic rewards inserted between sparse extrinsic rewards. In this approach, the intrinsic reward signal is not related to the extrinsic sparse reward and the final outcome of the game is not credited to each of the agent's behaviors. The lack of credit assignment may affect learning efficiency, especially when there is conflict between the agent's short-term and long-term goals, as is the case in our setup.

Among learning algorithms for distributed decision making, no-regret algorithms apply to a wide range of problems and converge fast; however, they require the knowledge of best strategies that are typically assumed to be static \cite{chang2007no}. Best-response algorithms search for best responses to other users' strategies, not for an equilibrium -- they therefore adapt to a dynamic environment, but they may not converge at all \cite{weinberg2004best}. To improve the convergence property of best-response algorithms, \cite{bowling2002multiagent} introduces an algorithm with varying learning rate depending on the reward; \cite{weinberg2004best} extends the work to non-stationary environments. However, both these algorithms provably converge only with restricted classes of games, and they are hard to implement in large or continuous state-action space, as is also the case in our set up of a multi-agent dynamic environment. The FSP method, on the other hand, addresses strategic agents' adaptiveness in a dynamic environment by incrementally evaluating state information and by keeping a weighted historical record \cite{heinrich2015fictitious}, and it is easy to implement in a large state space. It therefore befits our requirements.

\begin{table}[t]
 \centering
 \captionof{table}{Sec.\ref{sec:modelproblem} symbol definition}
 \label{tab:problem}
 \begin{tabular}{c l c l c l c }
 Sym & Description & Sym & Description\\
 \toprule
 $m \in M$ & bidder & $v$ & bid value\\
 $\alpha$ & backoff decision & $b$ & bidding price\\
 $c$ & joining cost & $q$ & backoff cost\\
 $p$ & payment & $z$ & bidding outcome\\
 $u$ & imediate payoff & $U$ & cumulated payoff\\
 \bottomrule 
 \end{tabular}
\end{table} 

\section{Problem Formulation}
\label{sec:modelproblem}

We formulate the long-term reward maximization problem in a generalized repeated auction that can be first- or second-price, forward or reverse, with any customized winning rules and payment scheme. Table \ref{tab:problem} summarizes the notation. 

Let $M$ be the set of bidders. Bidder $m \in M$ has at most $1$ demand for the commodity at $t$, denoted as $m^t \in \{0,1\}$. Bidder $m$ has two actions: whether to back off $\mathbf{\alpha}_m^t \in \{0,1\}$, and which price to bid $b_m^t \in \mathbb R_+$; the bidder draws them from a strategy. Bidder $m$ determines its bidding price $b_m^t$ using some function $f_m$ of its private valuation $v_{m} \in \mathbb R_+$ of the commodity, hence $b_{m}^t=f_m(v_{m})$; $f_m$ is only known to the bidder. The competing bidders draw their actions from a joint distribution $\pi_{-m}^t$ based on $(p^1,\cdots,p^{t-1})$, where $p^t \in \mathbb R_+$ is the final price of the commodity at time $t$. The final price is a function of all bidding prices at time $t$: $p^t=g(\mathbf{b}^t),\mathbf{b}^t\in \mathbb R_+^{|M|}$, depending on the auction mechanism we use; for example, in a first price lowest-bid auction, $g(\cdot)=\text{min}(\cdot)$. Bidder $m$ observes the new $p^t$ as feedback. The commodity is granted to the bidder with the highest score according to the broker's internal logic, for example, in any lowest-bid auction, bidder $m$'s score is a linearly decreasing function of $b_m^t$. $m$'s utility is denoted by $u_m(b_m^t,z_m^t), z_m^t \in \{0,1\}$, if $z_m^t=1$, $m$ wins. A winning bidder receives an immediate payoff that is a function of $m$'s private value $v_m^t$ of the commodity, its bidding price $b_m^t$, and the final price $p^t$; loosing bidders receive a negative payoff $c_m^t$ as cost to join the auction, and bidders that backed off receive a negative payoff $q_m^t$ as cost of backoff. We write the payoff as $u_m^t=h(v_m^t,b_m^t,z_m^t,p^t,c_m^t,q_m^t)$. The auction repeats for $T$ periods. Each bidder's goal is to independently maximize its long-term utility: $\mathcal{U}= \frac{1}{T} \sum\limits_{t=1}^T u_m(b_m^t), T\to \infty$.

\begin{table}[t]
 \centering
 \captionof{table}{Sec.\ref{subsec:greedy} symbol definition}
 \label{tab:greedy}
 \begin{tabular}{c l c l c l c }
 Sym & Description & Sym & Description\\
 \toprule
 $\zeta$ & best response & $\psi$ & behavioral strategy\\
 $\mathbf e_m$ & env. variables & $\rho$ & private bidder info\\
 $\mathbf{a}$ & action, $\mathbf{a}=(\alpha, b)$ & $P_{-m}^t$ & other bidders state\\
 $\text{sl}_m^t$ & SL present state & $\text{rl}_m^t$ & RL present state\\
 $S_m^t$ & RL complete state & $\lambda$ & $\bar{u}$'s weight factor\\
 $\theta$ & actor parameters & $\mathbf w$ & critic parameters\\
 $\gamma$ & learning rate & $\delta$ & TD error\\
 $\eta$ & $\zeta$'s weight & $\nu$ & history length\\
 $\mu$ & action mean & $\Sigma$ & action covariance\\
 \bottomrule 
 \end{tabular}
\end{table}

\begin{figure}[t]
	\centering
	\includegraphics[width=0.9\linewidth]{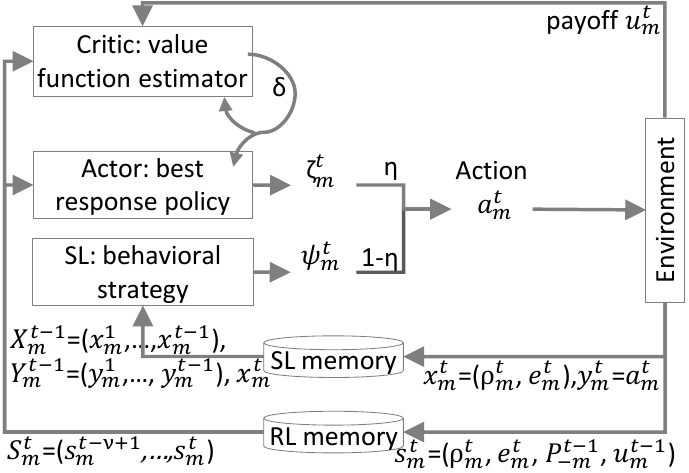}
	\caption{Short-term algorithm}
	\label{fspchart}
\end{figure}

\section{Proposed Solution}
\label{sec:solution}

To solve the long-term reward maximization problem described in Sec.~\ref{sec:modelproblem}, we propose an RL algorithm for long-term reward maximization. We first introduce the benchmark short-term algorithm in Section~\ref{subsec:greedy}; that algorithm maximizes a short-term reward. Then, in Section~\ref{subsec:longterm}, we introduce our long-term reward maximization algorithm that is based on the short-term algorithm.

\subsection{Short-Term Algorithm}
\label{subsec:greedy}

	  \begin{algorithm}[t]
	  \small
	  \begin{algorithmic}[1]
	  \STATE Initialize $\psi_m,\zeta_m$ arbitrarily, $t=1,\eta=1/t,\nu,P_{-m}^{t-1}=\mathbf{0},u_m^{t-1}=0$, observe $e_m^t$, create $\text{rl}_m^t,\text{sl}_m^t$ and add to memory
	  \WHILE{true}
	    \STATE Take action $\mathbf{a}_m^t=(1-\eta)\psi_m^t+\eta \zeta_m^t$
	    \STATE Receive $P_{m}^t$, calculate $u_m^t$, observe $\rho_m^{t+1},\mathbf e_m^{t+1}$
	    \STATE Create and add state to RL memory: $\text{rl}_m^{t+1}$
	    \STATE Create and add state to SL memory: $(\text{sl}_m^{t+1},\mathbf{a}_m^t)$
	    \STATE Construct $S_m^t,S_m^{t+1}$, calculate $\zeta_m^{t+1}=\text{RL}(S_m^t,S_m^{t+1},u_m^t)$
	    \STATE Calculate $\psi_m^{t+1}=\text{SL}(\text{sl}_m^{t+1})$
	    \STATE $t \gets t+1$, $\eta \gets 1/t$
	  \ENDWHILE
	  \end{algorithmic}
	  \caption{FSP algorithm for bidder $m$}
	  \label{algorithm}
	  \end{algorithm}

	  \begin{algorithm}[t]
	  \small
	  \begin{algorithmic}[1]
	  \STATE Initialize $\theta, w$ arbitrarily. Initialize $\lambda$
	  \WHILE{true}
	    \STATE Input $t$ and $S_m^t,S_m^{t+1}$ constructed from RL memory
	    \STATE Run critic and get $\hat V(S_m^{t}, \mathbf w),\hat V(S_m^{t+1},\mathbf w)$
	    \STATE Calculate $\bar u_m=\lambda \bar u_m$ and $\delta$ (immediate payoff $u$ is reward)
	    \STATE Run actor and get $\mu(\theta), \Sigma(\theta)$
	    \STATE Sample $\zeta_m^{t+1}$ from $F(\mu,\Sigma)$, update $\mathbf w$ and $\theta$
	  \ENDWHILE
	  \end{algorithmic}
	  \caption{RL algorithm for bidder $m$}
	  \label{algorithmRL}
	  \end{algorithm}

The short-term algorithm is based on the FSP method, it addresses the convergence challenge of a best-response algorithm. FSP balances exploration and exploitation by replaying its own past actions to learn an average behavioral strategy regardless of other bidders' strategies; then it cautiously plays the behavioral strategy mixed with best response \cite{heinrich2015fictitious}. Table \ref{tab:greedy} summarizes the notation for this section. The method consists of two parts: a supervised learning (SL) algorithm predicts the bidder's own behavioral strategy $\psi$, and an RL algorithm predicts its best response $\zeta$ to other bidders. The bidder has $\eta,\lim\limits_{t \to \infty} \eta =0$ probability of choosing action $\mathbf{a}=\zeta$, otherwise it chooses $\mathbf{a}=\psi$. The action includes backoff decision $\alpha$ and bidding price $b$. If $\alpha$ is above a threshold, the bidder submits the bid; otherwise, the bidder backs off for a duration linear in $\alpha$. We predefine the threshold to influence bidder behavior: with a higher threshold, the algorithm becomes more conservative and tends to back off more bids. Learning this threshold (e.g.\ through meta-learning algorithms) is left for future work.

Although FSP only converges in certain classes of games \cite{LESLIE2006285} -- and in our case of a multi-player, general-sum game with infinite strategies, it does not necessarily converge to an NE -- it is still an important experiment as our application belongs to a very general class of games; empirical results show that by applying FSP, overall performance is greatly improved compared to using only RL. The FSP is described in Alg.~\ref{algorithm}. 

Input to SL includes bidder $m$'s current bidder information $\rho_m^t$ (e.g.\ initial conditions, current reserve pool, etc.), and environment information visible to $m$, denoted $e_m^{t}$ (e.g.\ number of bidders in the network, number of active bids, final price in the previous round, etc.). SL infers behavioral strategy $\psi_m^t$. The input $\text{sl}_m^t=(\rho_m^t,e_m^t)$ and actual action $\mathbf{a}_m^t$ are stored in SL memory to train the regression model. We use a multilayer perceptron in our implementation. 

Input to RL is constructed from $m$'s present state $\text{rl}_m^t$. $\text{rl}_m^t$ is a combination of $\rho_m^t$; $e_m^t$; previous other bidders' state $P_{-m}^{t-1}$, represented by the final price $p$, or $P_{-m}^t=\mathbf{p}^t=\{ p_k^t| k \in K \}$; and calculated immediate payoff $u_m^{t-1}$. To consider historical records, we take $\nu$ most recent states to form the complete state input to RL: $S_m^t=\{\text{rl}_m^\tau|\tau=t-\nu+1,\cdots,t\}$. RL outputs best response $\zeta_m$ (Fig.~\ref{fspchart}). We provide a detailed description of the RL algorithm below.

\subsubsection{The RL Algorithm}
\label{modelDescription}

Our approach is similar to \cite{khaledi2016optimal} in the use of a learning algorithm for the bidders to adjust their bidding price based on budget and observation of other bidders: we estimate other bidders' state $P_{-m}$ from payment information and use the estimate as basis for a policy. Also, similar to their work, payment information is only from the broker to each bidder. However, our approach differs from \cite{khaledi2016optimal} in several major points. We use a continuous space for bidder states (i.e.\ continuous value for payments). As also mentioned in \cite{khaledi2016optimal}, a finer-grained state space yields better learning results. We do not explicitly learn the transition probability of bidder states. Instead, we use historical states as input and directly determine the bidder's next action.

\begin{table}[t]
 \centering
 \captionof{table}{Sec.\ref{subsec:longterm} symbol definition}
 \label{tab:longterm}
 \begin{tabular}{c l c l c l c }
 Sym & Description & Sym & Description\\
 \toprule
 $\theta$ & featurized state & $\epsilon$ & credit assign. weight\\
 $r_{i}$ & intrinsic reward & $r_{e}$ & extrinsic reward\\
 $L_{f}$ & forward mdl loss & $L_{i}$ & inverse mdl loss\\
 $\xi$ & reward weight\\
 \bottomrule 
 \end{tabular}
\end{table}
\begin{figure}[t]
	\centering
	\includegraphics[width=0.9\linewidth]{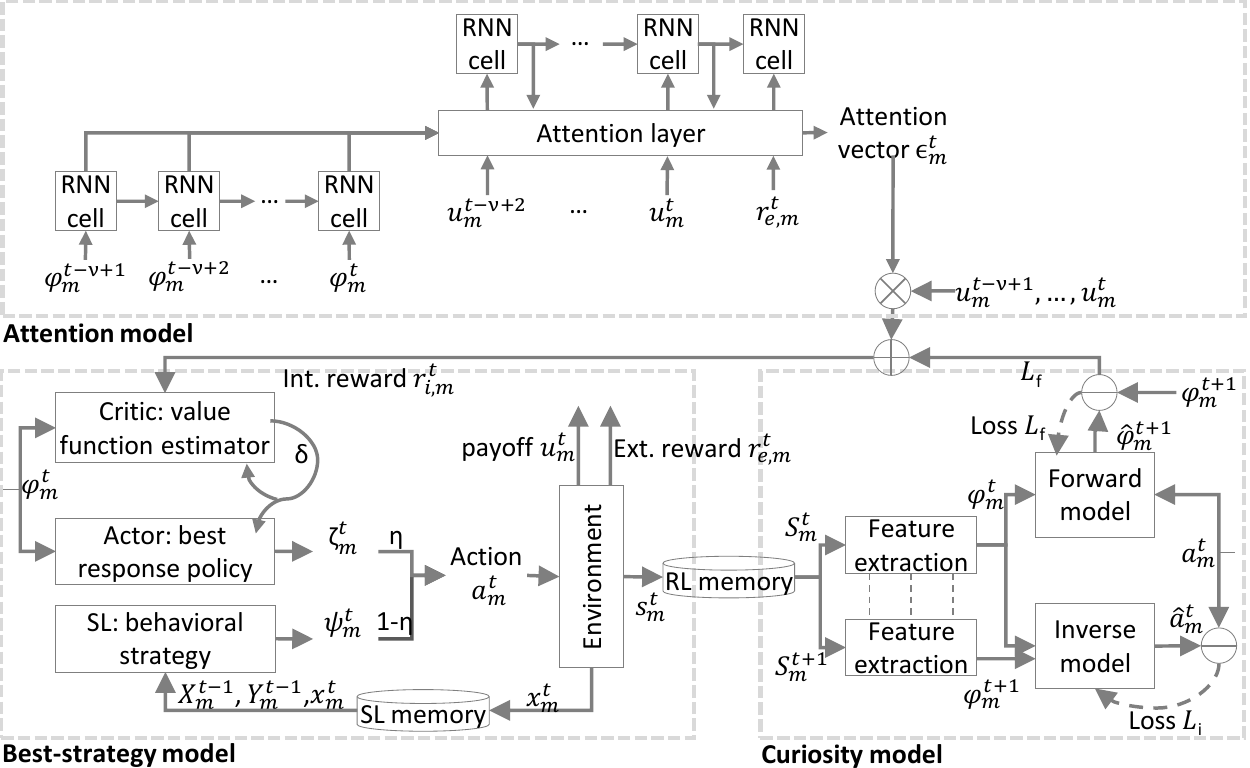}
	\caption{Long-term algorithms}
	\label{attentionchart}
\end{figure}

We use the actor-critic algorithm \cite{sutton2018reinforcement} for RL (Fig.~\ref{fspchart} and Alg.~\ref{algorithmRL}). The \textbf{critic} learns a state-value function $V(S)$. Parameters of the function are learned through a neural network that updates with $\mathbf w \gets \mathbf w + \gamma^w\delta \nabla \hat V(S, \mathbf w)$, where $\gamma$ is the learning rate and $\delta$ is the TD error. For a continuing task with no terminal state, no discount is directly used to calculate $\delta$. Instead, the average reward is used \cite{sutton2018reinforcement}: $\delta =u-\bar u+\hat V(S',\mathbf w) - \hat V(S,\mathbf w)$. In our case, the reward is utility $u$. We use exponential moving average (with rate $\lambda$) of past rewards as $\bar u$.

The \textbf{actor} learns the parameters of the policy $\pi$ in a multidimensional and continuous action space. Correlated backoff and bidding price policies are assumed to be normally distributed: $F(\mu,\Sigma) = \frac{1}{\sqrt{|\Sigma|}} \exp(-\frac{1}{2}(\mathbf x-\mu)^T\Sigma^{-1}(\mathbf x - \mu))$. For faster calculation, instead of covariance $\Sigma$, we estimate lower triangular matrix $L$ ($LL^T=\Sigma$). Specifically, the actor model outputs the mean vector $\mu$ and the elements of $L$. Actor's final output $\mathbf{\zeta}$ is sampled from $F$ through $\mathbf{\zeta} = \mu + L\mathbf{y}$, where $\mu$ is the mean and $\mathbf{y}$ is an independent random variable from standard normal distribution. Update function is $\theta \gets \theta + \gamma^\theta \delta \nabla \ln \pi(\mathbf{a}|S,\theta)$. We use $\frac{\partial \ln F}{\partial \mu} =\Sigma (\mathbf x-\mu)$ and $\frac{\partial \ln F}{\partial \Sigma} = \frac{1}{2} (\Sigma(\mathbf x-\mu)(\mathbf x-\mu)^T\Sigma-\Sigma)$ for back-propagation.

The objective is to find a strategy that, given input $S_m^t$, determines $\mathbf{a}$ to maximize $\frac{1}{T-t}\mathbb{E}[\sum_{t'=t}^T u_m^{t'}]$. To implement the actor-critic RL, we use a stacked convolutional neural network (CNN) with highway \cite{srivastava2015training} structure similar to the discriminator in \cite{yu2017seqgan} for both actor and critic models. The stacked CNN has diverse filter widths to cover different lengths of history and extract features, and it is easily parallelizable, compared to other sequential networks. Since state information is temporally correlated, such a sequential network extracts features better than multilayer perceptrons. The highway structure directs information flow by learning the weights of direct input and performing non-linear transform of the input.


	  \begin{algorithm}[t]
	  \small
	  \begin{algorithmic}[1]
	  \STATE Initialize model parameters and $\epsilon$ arbitrarily. Initialize $\xi$
	  \WHILE{true}
	    \STATE Input $t$ and $S_m^t,S_m^{t+1}$ constructed from RL memory
	    \STATE Run feature extraction and get feature vector $\phi_m^t$ and $\phi_m^{t+1}$
	    \STATE Get actual action $a_m^t$ from the FSP algorithm
	    \STATE Run forward model, get $\hat \phi_m^{t+1}$, calculate $L_f$
	    \STATE Run inverse model, get $\hat a_m^t$, calculate $L_i$
	    \STATE Update forward, inverse, feature extraction model params
	    \STATE Infer from credit assignment model and get $\epsilon$
	    \STATE Calculate and output $r_{i,m}^t$
	  \ENDWHILE
	  \end{algorithmic}
	  \caption{Long-term learning algorithm}
	  \label{algorithmcuriosity}
	  \end{algorithm}

\subsection{Long-Term Learning Algorithm}
\label{subsec:longterm}

Our core contribution is the long-term reward maximization algorithm called DRACO2. It is based on the short-term RL algorithm from the previous section. We add the following features: 1) reward prediction, 2) more exploration in the early stages of learning, and 3) short- and long-term reward alignment through credit assignment. Points 1 and 2 are achieved through an adapted curiosity model. Point 3 is achieved through a hierarchical structure. The structure uses an attentional network that is responsible for learning to assign weights to short-term rewards based on their relevance to the long-term, sparse extrinsic reward, the learning process is only triggered when a new extrinsic reward becomes available (Fig.~\ref{attentionchart}). Between the extrinsic reward signals, the underlying RL+curiosity model learns to better predict next states, actions, and intrinsic rewards. Table \ref{tab:longterm} summarizes the notation for this section. Next, we describe the curiosity learning and credit assignment models in detail.

\subsubsection{Curiosity Model}
\label{curiosity}

Our curiosity model is based on the vanilla model from \cite{pathak2017curiosity}. The original model uses a feature extraction model to identify features that can be influenced by the agent's actions, thus improving the model's generalization properties in new environments. In our competitive environment, this ability is very important. Due to the nature of a repeated auction, next state depends not only on the current state and action, but on a number of historical state-actions. We therefore extract features from $S_m^t$. The resulting feature vector $\phi_m^t=\text{feature}(S_m^t)$ replaces $S_m^t$ in the previous short-term algorithm, to become the input of both the actor and the critic models. 

The original curiosity model uses a forward model and an inverse model to predict next state and next action, respectively. These are supervised learning models with the objective to minimize loss $L_f=\| \phi_m^t-\hat\phi_m^t \|_2^2$ and $L_i=\| \mathbf{a}_m^t-\hat{\mathbf{a}}_m^t \|_2^2$. One of the objectives of the forward and inverse models is to improve prediction accuracy of the consequence of the agent's actions, even without any reward signal. In our game setup, we have short-term intrinsic reward signals (only not aligned and potentially conflicting with the extrinsic rewards); therefore, we adapt the input to include the previous intrinsic reward values, and the forward model's objective is to improve prediction accuracy of both the state and the intrinsic reward. 

In the original curiosity model, the intrinsic reward is the loss of the forward model: $r_{i,m}^t=\xi L_f$, and the bigger the forward loss, the higher the intrinsic reward. Through the adversarial design, the model is encouraged to explore state-actions where the agent has less experience and prediction accuracy is low. The intrinsic rewards are inserted between sparse extrinsic rewards to improve learning efficiency despite the sparseness -- the authors of \cite{pathak2017curiosity} call this internal motivation ``curiosity-driven exploration''. In our approach, we apply the same method with a modified intrinsic reward definition: $r_{i,m}^t=\xi L_f^t + (1-\xi) \epsilon u_{m}^t$, where $\xi$ is a pre-defined weight factor to balance between the two short-term objectives, and $\epsilon$ is a weight factor from the credit assignment model (see below). The objective is to maximize: $\mathbb{E}_\pi [\sum_t r_{i,m}^t]-L_i-L_f$. The modified long-term learning algorithm based on the short-term algorithm is in Alg.~\ref{algorithmcuriosity}. Note that the FSP and Actor-Critic parts are the same as in Alg.~\ref{algorithm} and \ref{algorithmRL}, except the input to both actor and critic is the feature vector $\phi_m^t$, instead of the original state vector $S_m^t$, and reward is $r_{i,m}^t$ instead of $u_m^t$.

	  \begin{algorithm}[t]
	  \small
	  \begin{algorithmic}[1]
	  \STATE Initialize model parameters arbitrarily, initialize batch size $n$
        \STATE Input $t$ and $S_m^t,\cdots,S_m^{t-n}$ from RL memory
	  \STATE Run feature extraction and get $\phi_m^t,\cdots,\phi_m^{t-n}$
	  \FOR{$i \gets t-n$ to $t$}
	    \STATE Input $\phi_m^i$ to encoder, run and get encoder output $e^i$
	    \STATE Input $e^i$ and target $u_m^i$ to decoder
	    \STATE Run decoder with attention layer
	  \ENDFOR
	  \STATE Extract weight vector $\epsilon_m^t$ from attention layer
	  \end{algorithmic}
	  \caption{Credit assignment algorithm}
	  \label{algorithmattention}
	  \end{algorithm}

\subsubsection{Credit Assignment Model}
\label{creditassign} 

The credit assignment model uses a sequential network (recurrent neural network as encoder and decoder) with an attention layer. Typically, such a sequential network is used to identify correlation between sequenced input elements $\text{enc}_i$ and predict a corresponding sequence of output elements $\hat{\text{dec}}_o$. The sequential network can be enhanced with an attention layer. In each time step, for every decoder output, the attention layer generates a weight vector corresponding to each input element, marking its relevance to the current output prediction. Model parameters are updated with the mean square error between the generated output $\hat{\text{dec}}_o$ and target vector $\text{dec}_o$.

In our credit assignment model, we are not interested in decoder output. Instead, we are interested in the contribution of each state-action pair towards the final extrinsic reward $r_{e,m}^t$. Therefore, we trigger the training of the credit assignment model only when there is a new signal $r_{e,m}^t$ at time $t$. This signal becomes the last element of the target vector. We train the model on the batch of $\nu$ featurized state vectors $\text{enc}_i=\{\phi_m^{t-\nu+1},\cdots,\phi_m^{t}\}$ with both short- and long-term rewards as target vector, $\text{dec}_o = \{u_{m}^{t-\nu+1},\cdots,u_{m}^t,r_{e,m}^t\}$, until the attention layer outputs a weight vector $\epsilon_m^t=\{\epsilon_1,\cdots,\epsilon_\nu|\sum_{i=1}^n \epsilon_i=1\}$ corresponding to $\text{enc}_i$ that marks their relevance to the last output $r_{e,m}^t$.

The weight vector $\epsilon_m^t$ is then multiplied with the original auction payoffs $u_m^t$. Through $\epsilon_m^t$, short- and long-term rewards are aligned, even if they are conflicting in nature. Between sparse extrinsic rewards, only the forward network of credit assignment model is run to infer a weight vector.

\section{Evaluation}
\label{sec:eval}

We train DRACO2 in two repeated auction games with a Python discrete event simulator. Both games have six bidder agents and one broker agent. The broker is a passive agent without learning capabilities. In every time step, the broker offers one commodity (e.g.\ object or service) for bidding, all bidders can join the auction simultaneously. The broker grants the commodity to the bidder with the highest score; ties are broken randomly. An immediate payoff is given to the winner, the value of the payoff is specific to the type of game. Except the winner, all other participating bidders pay a fixed cost for joining the auction.

\begin{figure*}[t]
	\centering
	\subcaptionbox{Payoff-performance per agent, by algorithm type.\label{fp_reward_het}}{\includegraphics[width=0.24\linewidth]{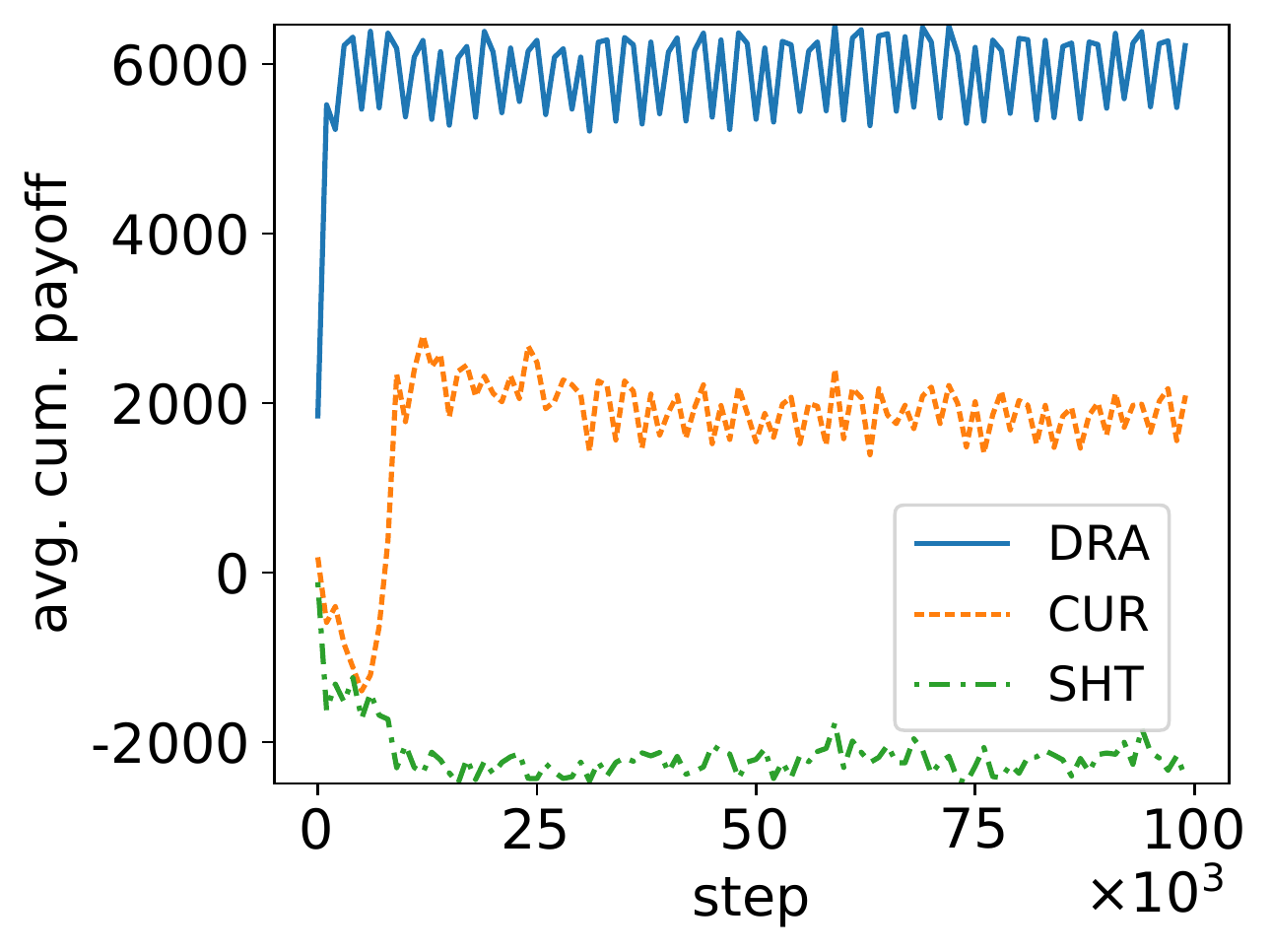}}\hfill
	\subcaptionbox{Fairness-performance.\label{fp_reward_het_Jindex}}{\includegraphics[width=0.23\linewidth]{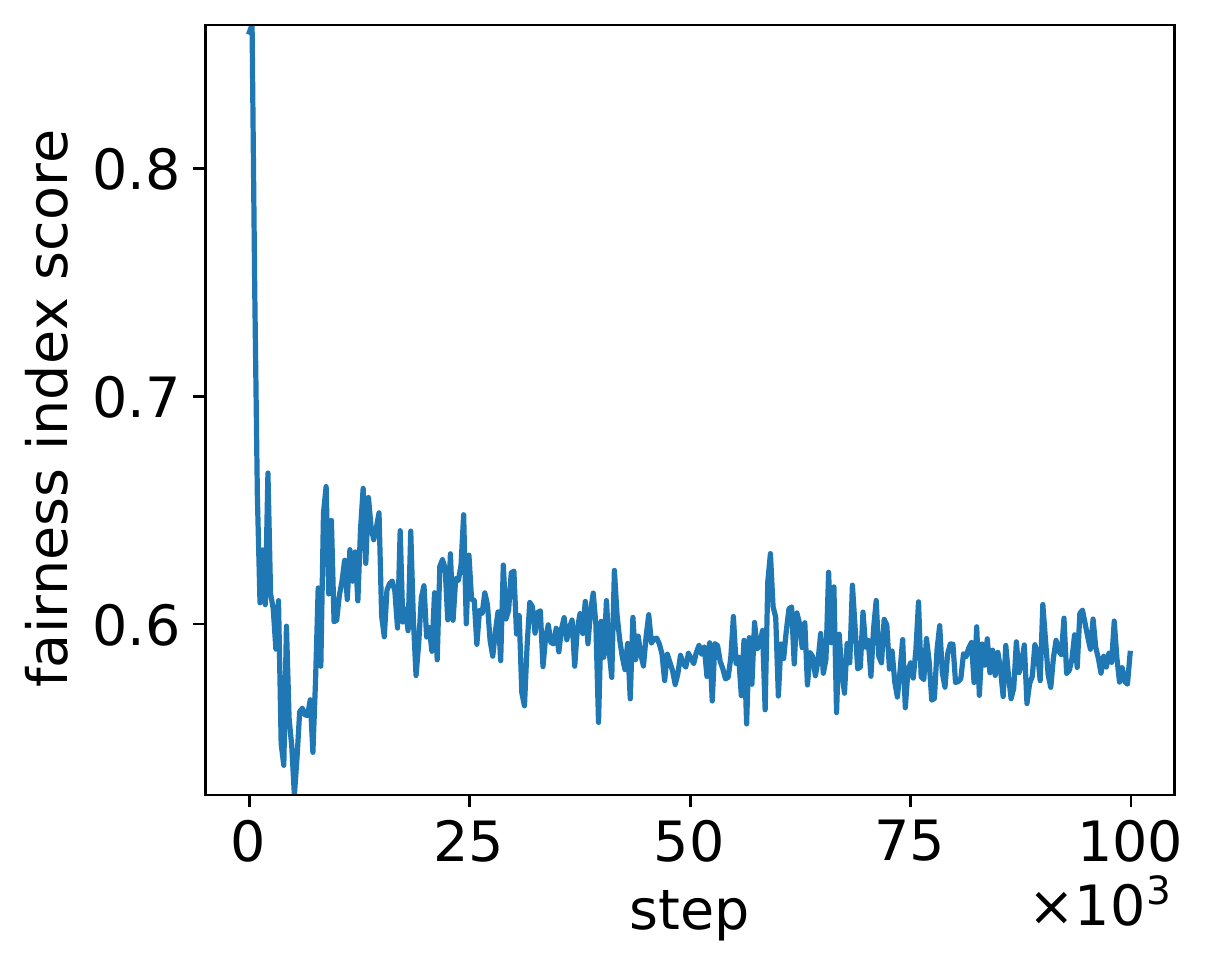}}\hfill
	\subcaptionbox{Intrinsic reward, comparison of DRA and CUR agents.\label{fp_het_intrwd}}{\includegraphics[width=0.23\linewidth]{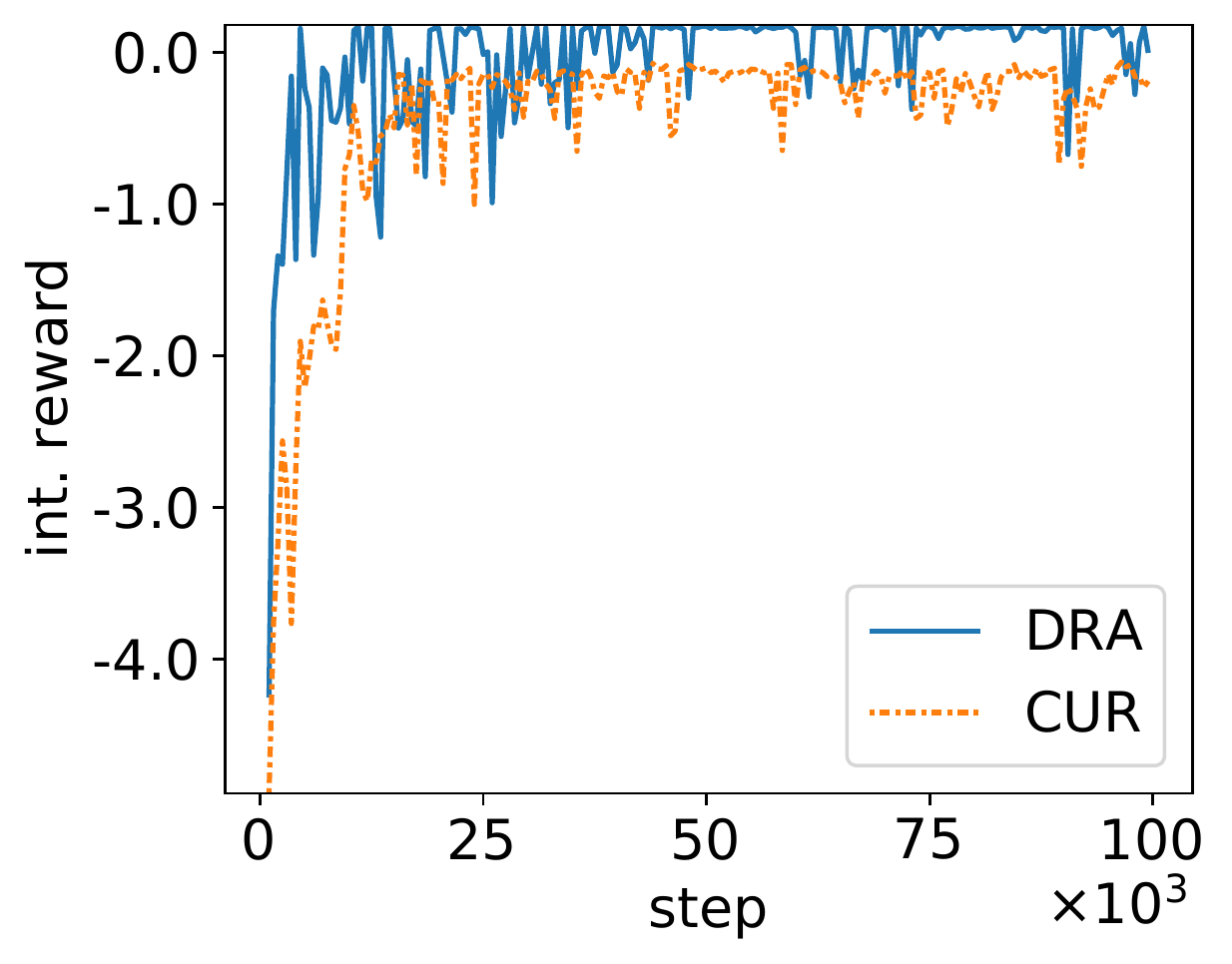}}\hfill
	\subcaptionbox{Forward model loss, comparison of DRA and CUR agents.\label{fp_het_fwdloss}}{\includegraphics[width=0.23\linewidth]{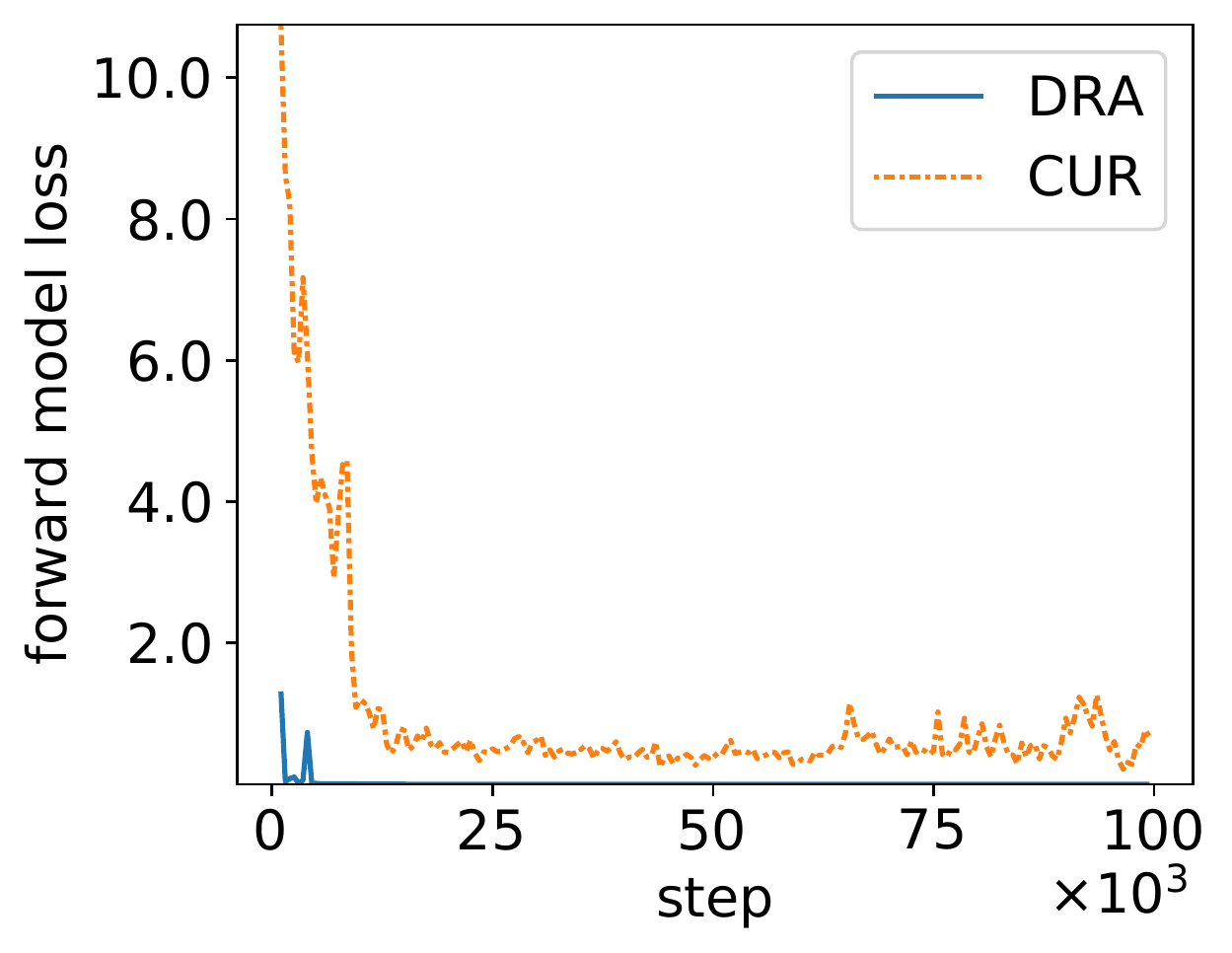}}\hfill
	\caption{FP with HETERO agents and payoff-signal}
	\label{fp_reward_het_perf}
\end{figure*}

\begin{figure}[t]
	\centering
	\subcaptionbox{Payoff-performance: 4/6 agents (bidders2-5) receive max. reward. \label{fp_reward_hom_att_peragent}}{\includegraphics[width=0.48\linewidth]{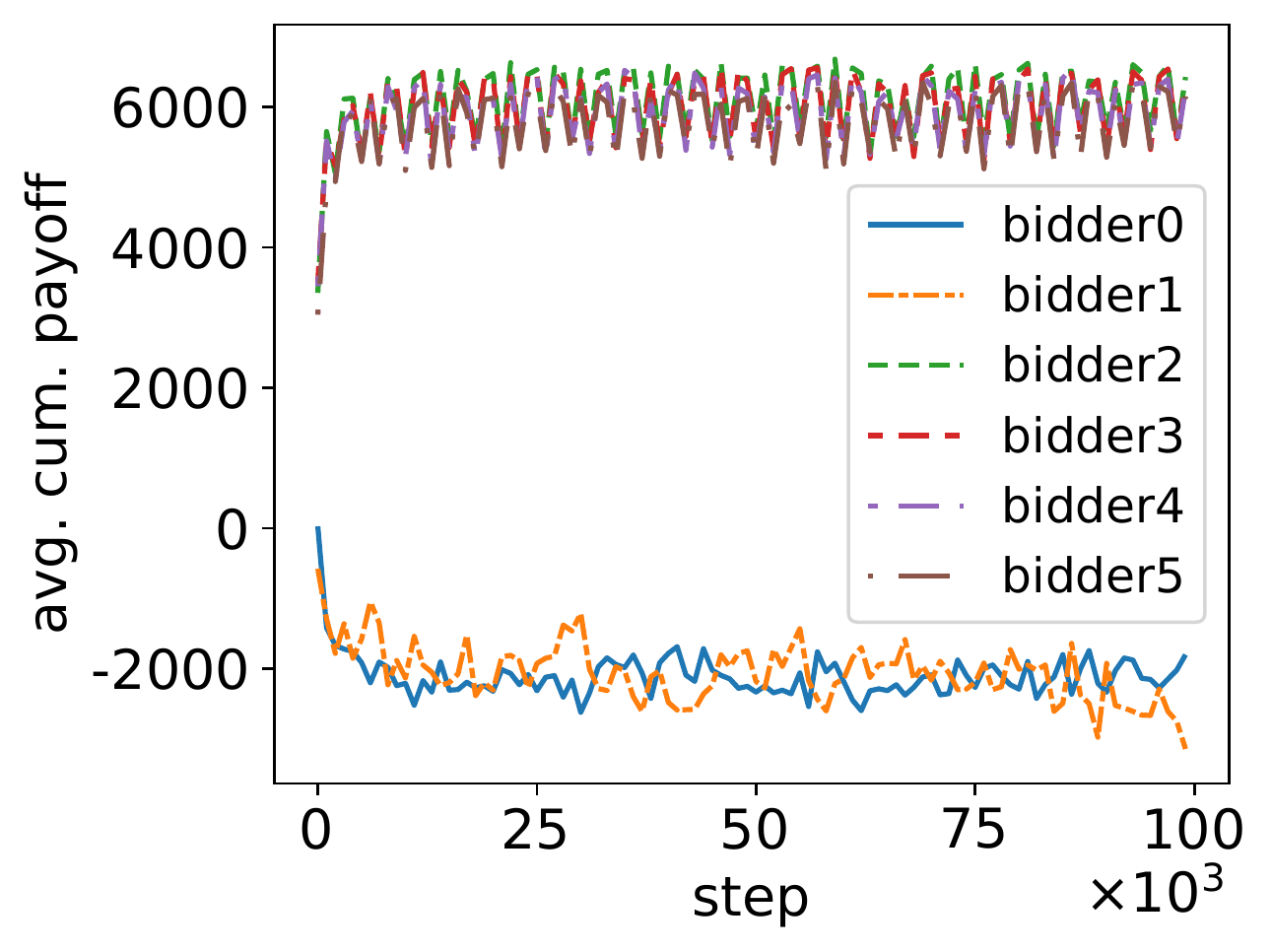}}\hfill
	\subcaptionbox{Payoff-performance comparison: all DRA vs. HETERO. \label{fp_reward_hom_att}}{\includegraphics[width=0.48\linewidth]{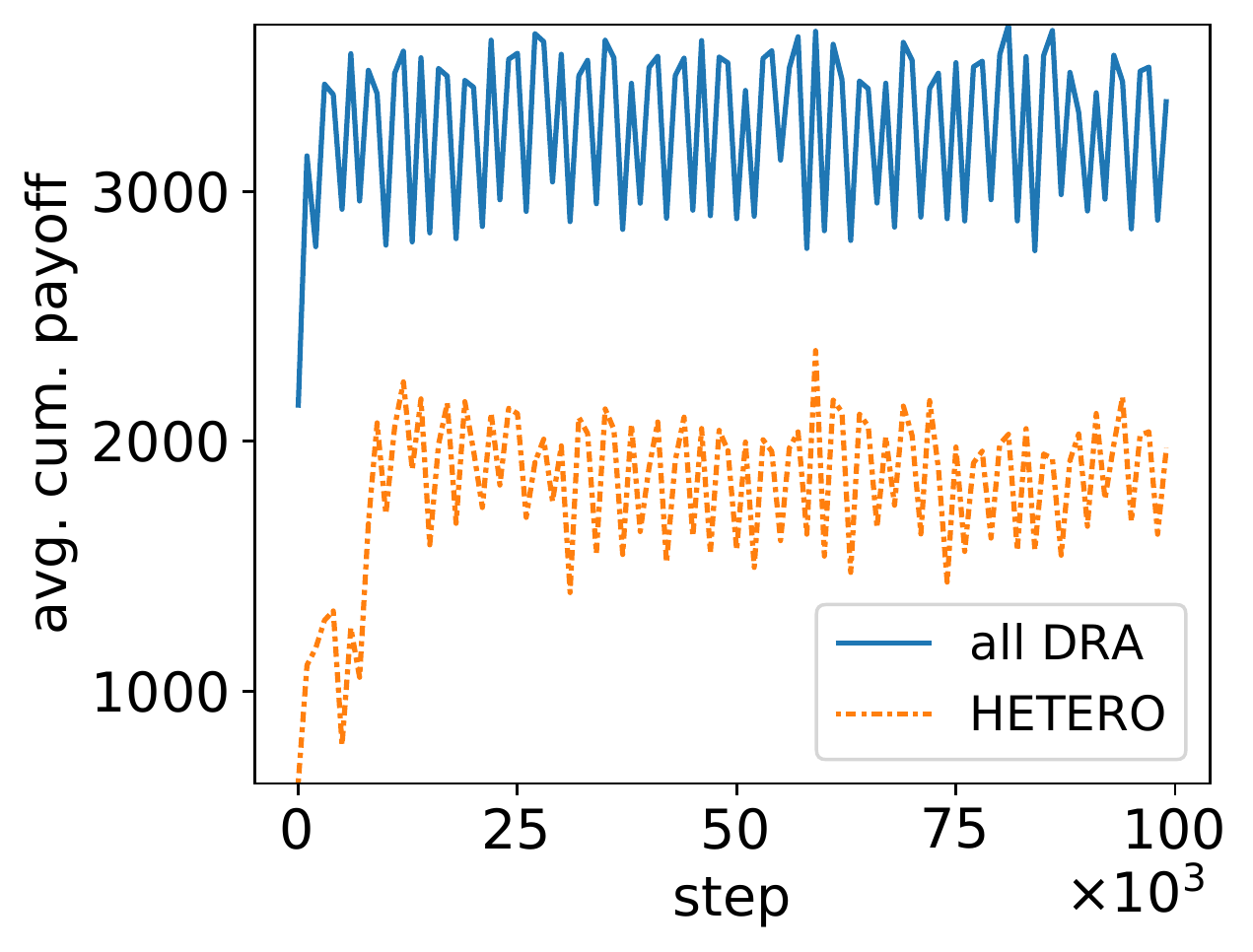}}\hfill
	\caption{FP with only DRA agents and payoff-signal}
	\label{fp_reward_hom}
\end{figure}

All bidders start with a reserve pool of wealth; it is updated every time step with payoffs and costs. Regardless of the bidder's behavior, there is a constant cost each time step (carrying cost). If the pool is depleted, the game is over for the bidder, it receives a penalty, and rejoins the game with the same initial reserve. Otherwise, the game continues for a certain number of time steps (in our simulation we take $T=150$ time steps). When the game ends, all bidders restart the game with the same initial reserve. In the case of long-term learning algorithms, bidder $m$ receives a long-term extrinsic reward signal at the end of each game. It can be $m$'s own cumulated payoff in the reserve pool of wealth: $r_{e,m}^t=U_m^t=\sum\limits_{t-T}^{t} u_m^\tau$, or overall fairness, defined as the J-index \cite{jain1984quantitative} of payments from the broker to the bidder agents over time: $r_{e,m}^t = \cfrac{(\sum\limits_{m}\sum\limits_{t-T}^t p_m^t)^2}{|M|\sum\limits_m (\sum\limits_{t-T}{t} p_m^t)^2},\forall m \in M$, which is commonly used to measure fairness in networking. J-index is the reciprocal of the original normalized Herfindahl–Hirschman Index \cite{rhoades1993herfindahl} used to measure market concentration. To preserve privacy, extrinsic reward signals do not contain private agent information.
There are free and occupied bidders: if a bidder wins a bid, its resources are occupied for a period of time, i.e.\ service duration, during which the occupied bidder cannot submit new bids. Each free bidder decides on 1) whether to join the auction for the commodity in the current time step, 2) if so, a bidding price $b$ that is lower than or equal to the amount in the reserve pool of wealth, and 3) other decision factors required by the specific auction setup. In the first-price reverse auction, service duration $d$ is correlated with the bidding price $b$. The broker gives bidders a balanced score based on the multiplication of price and service duration. 
Winner of the auction gets a immediate payoff of $b \cdot d$. In the second-price forward auction, bidders decide on bidding price $b$, and $d$ is a constant value (to simplify winning criterion and the calculation of second-price). If it wins, the bidder pays the broker the second-highest bidding price $p$ among all bidders. 
The winner gets an immediate payoff of $(b-p) \cdot d$. In both games, during the service duration $d$, the winner cannot join any new auctions.

The bidders may use one of three learning algorithms: the short-term algorithm (SHT), the long-term algorithm based on curiosity learning (CUR), and DRACO2, the long-term algorithm with an attention layer for credit assignment (DRA). In the setup with homogeneous agents, all six agents have an algorithm of the same type. In the setup with heterogeneous agents, each algorithm is given to two bidders, and all algorithms compete in the same auction game.

To summarize, we simulate first-price reverse (FP) or second-price forward (SP) auction, with homogeneous (DRA, CUR or SHT) or heterogeneous agents (HETERO), and use either average cumulated payoff per agent (payoff-signal), or fairness index score (fairness-signal) as extrinsic reward signals. As performance measure we measure the average cumulated payoff per agent (payoff performance), and the overall fairness index score (fairness performance). All results come from continuous training.

\subsection{First-Price Reverse Auction (FP)}

First-price reverse auction (lowest-bid-wins) is common e.g.\ in long-term energy contracts \cite{lucas2013renewable} or network resource allocation \cite{xu2012resource} where multiple resource owners bid to sell to one buyer that prefers low price for a long duration.

Each curve in Fig.~\ref{fp_reward_het} represents the average performance of two agents with the same type of algorithm, in a heterogeneous setting. Both DRA and CUR agents outperform SHT: through the reserve pool of wealth, current behavior influences bidding decisions in the future and has direct impact on the delayed extrinsic reward. However, the short-term algorithm values the immediate intrinsic reward much higher than the extrinsic reward in the distant future, therefore failing to compete in the game. On the other hand, the DRA agents clearly performs the best, but at the cost of other agents with less aggressive algorithms, as is shown by the low fairness index in Fig.~\ref{fp_reward_het_Jindex}. Figures~\ref{fp_het_intrwd} and \ref{fp_het_fwdloss} compare training performance of DRA and CUR agents in the game. The DRA agent not only converges faster, it also converges to a lower loss and higher intrinsic reward.

If we pitch the aggressive DRA agents against each other, i.e.\ all six agents are DRA agents, we have similar a result (Fig.~\ref{fp_reward_hom_att_peragent}): only four DRA agents can maximize their cumulated payoff over time, although the game has a higher social welfare, compared to the HETERO case (Fig.~\ref{fp_reward_hom_att}). The difference in individual performance is caused by DRA agents' aggressive, selfish (i.e.\ with private indiviual goals), rational (i.e.\ act to maximize reward) behavior. They profit from an unregulated system at the cost of social welfare. In fact, it is possible for all six agents to maximize their reward: to motivate cooperation, we replace the cumulated payoff with fairness index score as long-term extrinsic reward signal. The negative impact on social welfare can thus be prevented.

\begin{figure}[t]
	\centering
	\subcaptionbox{Payoff-performance comparison: payoff-signal vs. fairness-signal.\label{fp_reward_hom_fair}}{\includegraphics[width=0.48\linewidth]{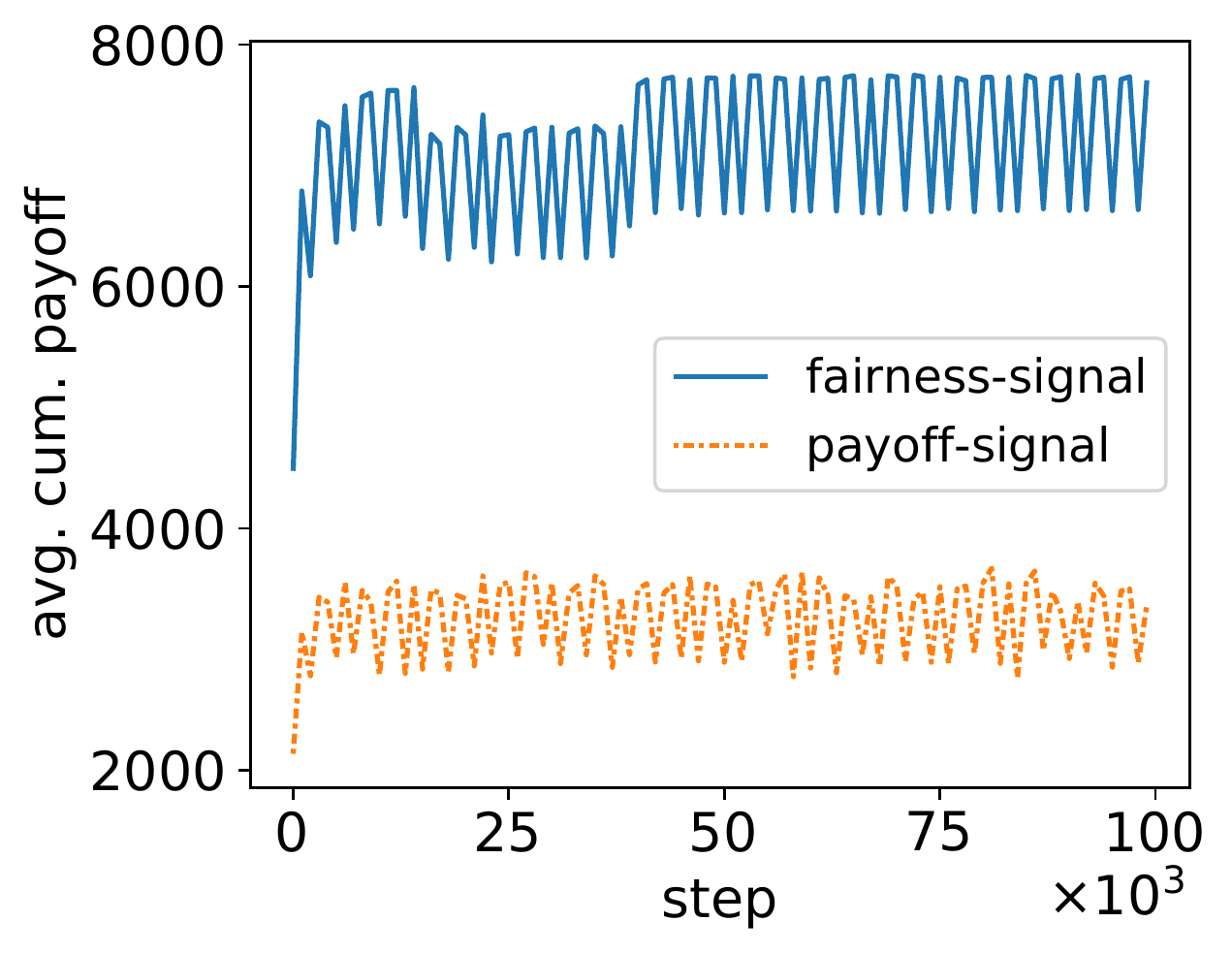}}\hfill
	\subcaptionbox{Fairness-performance comparison: payoff-signal vs. fairness-signal.\label{fp_reward_hom_Jindex}}{\includegraphics[width=0.47\linewidth]{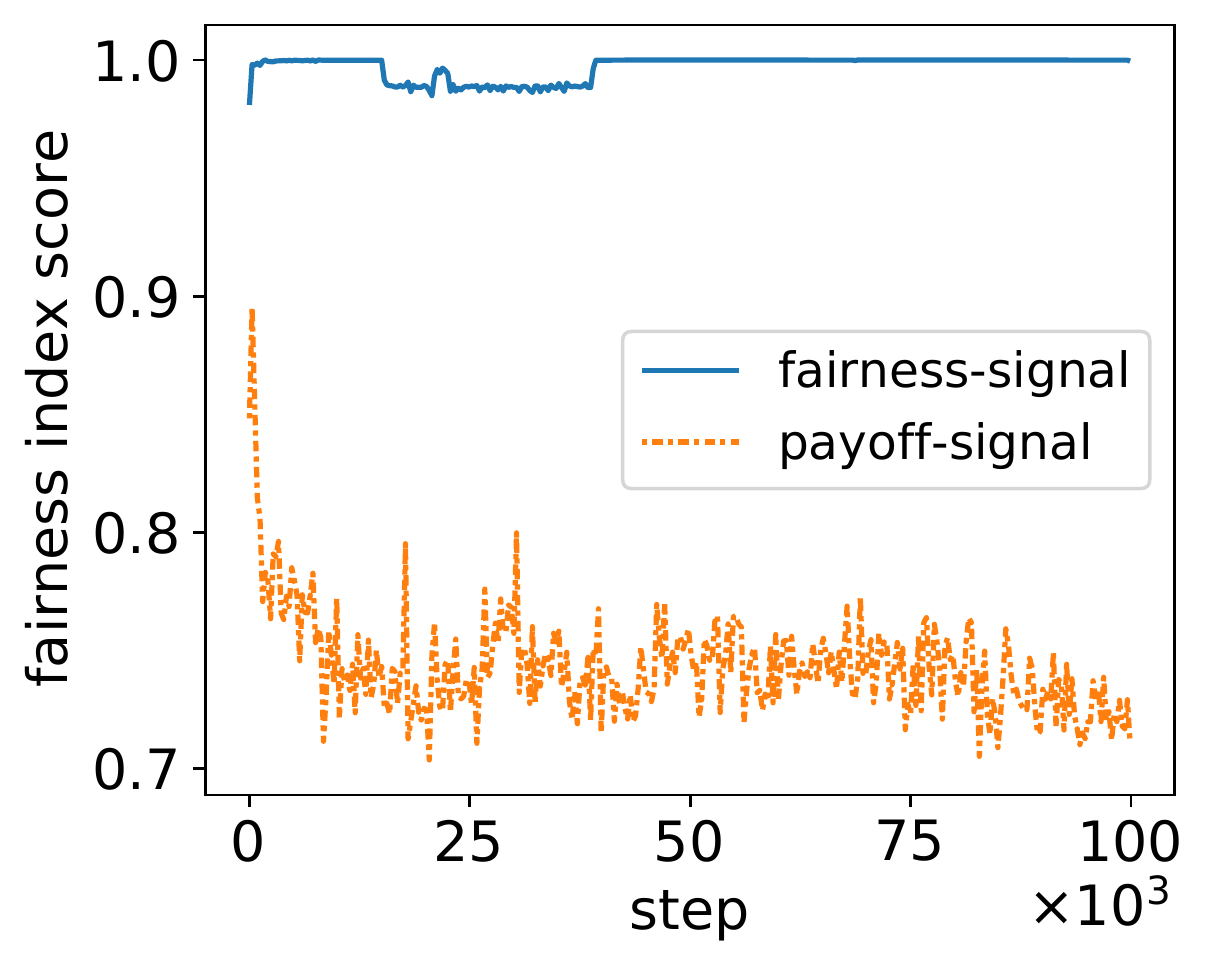}}\hfill	
	\caption{FP with only DRA agents and fairness-signal}
	\label{fp_rewardcomparison}
\end{figure}

Fig.~\ref{fp_rewardcomparison} compares two independent simulation results. The dotted orange curve is the average cumulative payoff of six DRA agents when the extrinsic reward is also the cumulative payoff. The solid blue curve is when the extrinsic reward is fairness index score. With fairness as incentive, all agents receive better cumulated payoffs, while achieving a much higher fairness index score. Hence, with our solution, it is possible to maximize both individual gain and social welfare.

To wrap up, the first simulation setup (Fig.~\ref{fp_reward_het_perf}) demonstrates how the DRACO2 algorithm learns quickly and aggressively in a multi-agent, dynamic environment with partial information, a big state-action space, and sparse / delayed extrinsic reward. The second setup (Fig.~\ref{fp_rewardcomparison}) demonstrates how DRACO2 can be easily optimized to integrate a system goal while preserving privacy and individual goals.

\subsection{Second-Price Forward Auction (SP)}

In a second-price forward auction (second-highest-bid-wins), the broker is a seller that grants the commodity to the bidder with highest bidding price, but the payment for the commodity is the second-highest price of all bidding prices. This type of auction is common for selling public goods, maximizing welfare rather than seller profit, e.g.\ in networking resource allocation \cite{xu2012interference} and e-commerce \cite{huang2011design}, where multiple end users bid for resources from one service provider.

\begin{figure}[t]
	\centering
	\subcaptionbox{Payoff-performance: HETERO w/ payoff-signal, by algorithm type.\label{sp_reward_het}}{\includegraphics[width=0.48\linewidth]{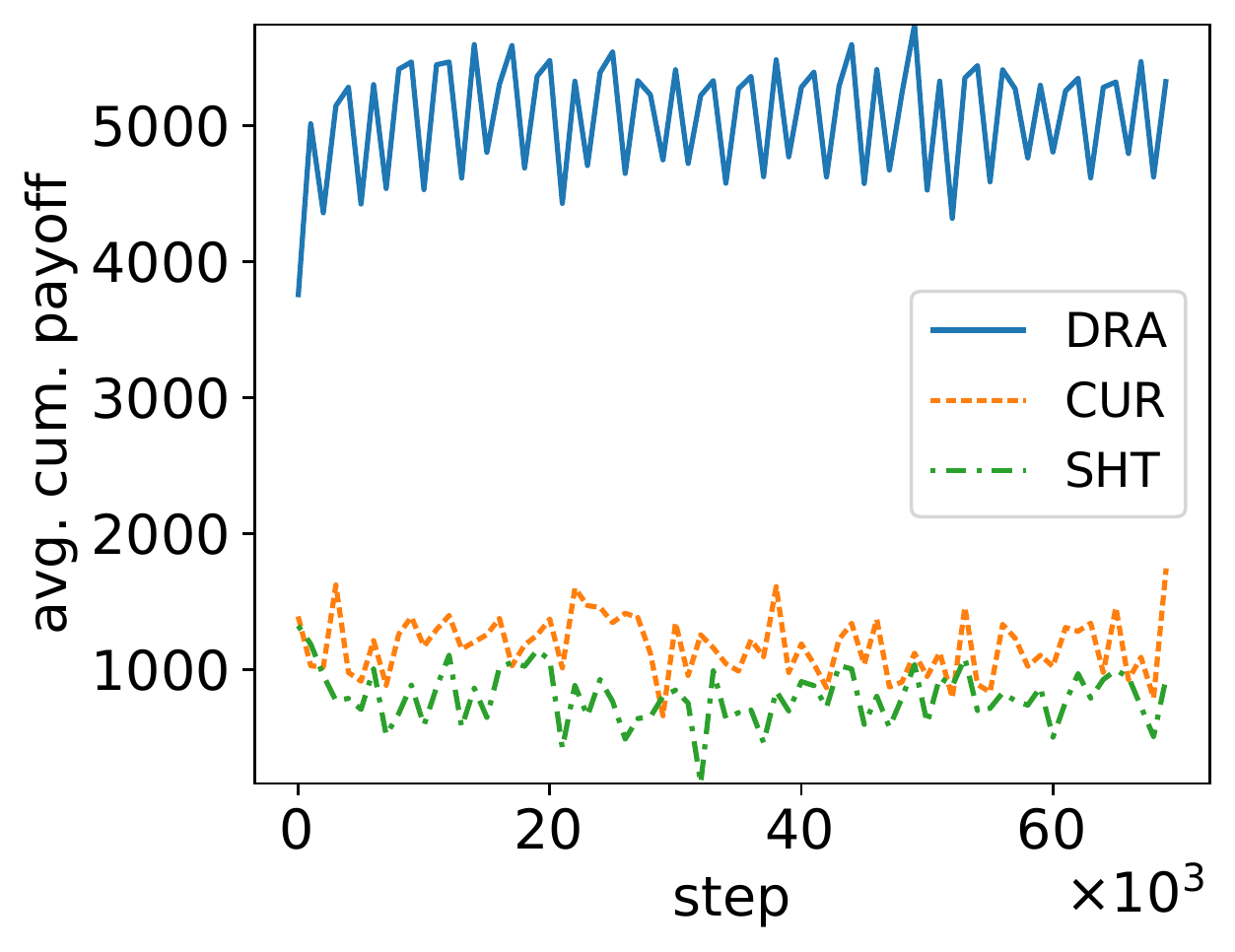}} \hfill
	\subcaptionbox{Payoff-performance comparison.\label{sp_reward_alltypes}}{\includegraphics[width=0.48\linewidth]{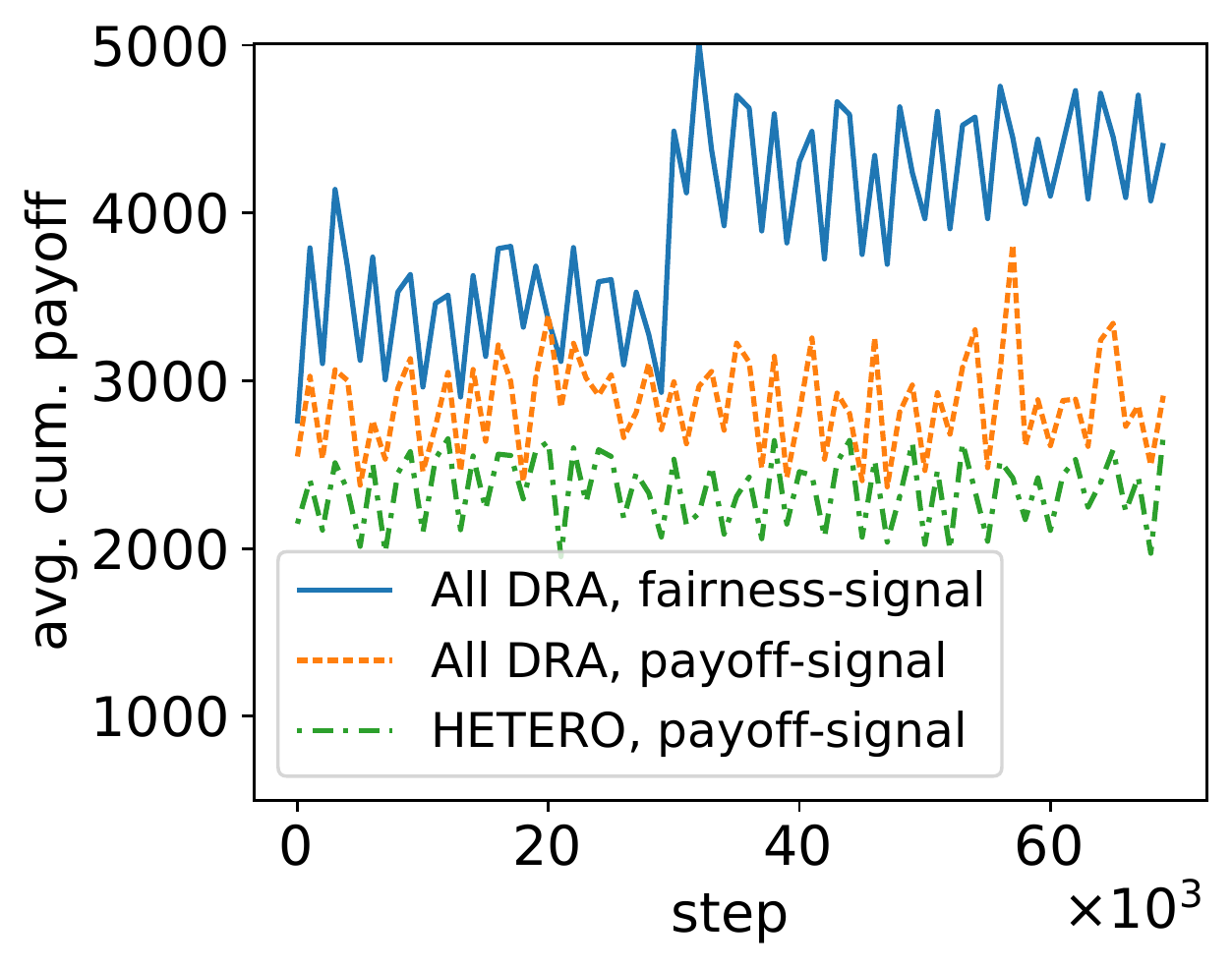}} \hfill
	\caption{SP, all DRA vs. HETERO, payoff-signal vs. fairness-signal}
	\label{sp_reward}
\end{figure}

Fig.~\ref{sp_reward} shows similar results in SP as in FP. When three types of agents co-exist in a profit-oriented setup, the two DRA agents win at the cost of social welfare (Fig.~\ref{sp_reward_het}). Social welfare increases when all agents are DRA agents. Finally, if the DRA agents are instead given a fairness index as incentive, social welfare reaches is much higher.
This can be seen from Fig.~\ref{sp_reward_alltypes}: with fairness index score as extrinsic reward signal, social welfare increases.

\section{Conclusion}
\label{sec:conclusion}

We demonstrate the performance of DRACO2 in two repeated auction games. The results show that, with the help of an attention layer for long-term credit assignment, the DRA agents behave more aggressively in the competition against other agents, when the long-term goal is to maximize cumulated private payoff. However, the selfish behavior has a negative impact on the overall social welfare. To encourage cooperation, we use fairness as the long-term goal. Simulation results show the improvement in individual payoff and in overall fairness index score.

We ran the simulations with only six agents, and the simulated auction mechanisms are relatively simple. In the next steps, we would focus on increasing the number of agents in the simulation, give them different individual goals, and test the algorithm in more complex setups, especially with more realistic extrinsic reward signals. 

\bibliographystyle{IEEEtran}
\bibliography{IEEEabrv,references}

\begin{thebibliography}{10}
\providecommand{\url}[1]{#1}
\csname url@samestyle\endcsname
\providecommand{\newblock}{\relax}
\providecommand{\bibinfo}[2]{#2}
\providecommand{\BIBentrySTDinterwordspacing}{\spaceskip=0pt\relax}
\providecommand{\BIBentryALTinterwordstretchfactor}{4}
\providecommand{\BIBentryALTinterwordspacing}{\spaceskip=\fontdimen2\font plus
\BIBentryALTinterwordstretchfactor\fontdimen3\font minus
  \fontdimen4\font\relax}
\providecommand{\BIBforeignlanguage}[2]{{%
\expandafter\ifx\csname l@#1\endcsname\relax
\typeout{** WARNING: IEEEtran.bst: No hyphenation pattern has been}%
\typeout{** loaded for the language `#1'. Using the pattern for}%
\typeout{** the default language instead.}%
\else
\language=\csname l@#1\endcsname
\fi
#2}}
\providecommand{\BIBdecl}{\relax}
\BIBdecl

\bibitem{xu2012resource}
C.~Xu \emph{et~al.}, ``Resource allocation using a reverse iterative
  combinatorial auction for device-to-device underlay cellular networks,'' in
  \emph{IEEE GLOBECOM}, 2012.

\bibitem{xu2012interference}
------, ``Interference-aware resource allocation for device-to-device
  communications as an underlay using sequential second price auction,'' in
  \emph{IEEE ICC}, 2012.

\bibitem{lucas2013renewable}
H.~Lucas \emph{et~al.}, ``Renewable energy auctions in developing countries,''
  \emph{International Renewable Energy Agency}, 2013.

\bibitem{huang2011design}
H.~Huang and R.~J. Kauffman, ``On the design of sponsored keyword advertising
  slot auctions: An analysis of a generalized second-price auction approach,''
  \emph{Electronic Commerce Research and Applications}, 2011.

\bibitem{schindler2011pricing}
R.~M. Schindler \emph{et~al.}, \emph{Pricing strategies: a marketing
  approach}.\hskip 1em plus 0.5em minus 0.4em\relax sage, 2011.

\bibitem{einav2018auctions}
L.~Einav \emph{et~al.}, ``Auctions versus posted prices in online markets,''
  \emph{Journal of Political Economy}, 2018.

\bibitem{busoniu2008comprehensive}
L.~Busoniu \emph{et~al.}, ``A comprehensive survey of multiagent reinforcement
  learning,'' \emph{IEEE Transactions on Systems, Man, and Cybernetics, Part C
  (Applications and Reviews)}, 2008.

\bibitem{teng2013reinforcement}
Y.~Teng \emph{et~al.}, ``Reinforcement-learning-based double auction design for
  dynamic spectrum access in cognitive radio networks,'' \emph{Wireless
  Personal Communications}, 2013.

\bibitem{almasri2020dynamic}
M.~Almasri \emph{et~al.}, ``Dynamic decision-making process in the
  opportunistic spectrum access,'' \emph{Advances in Science, Technology and
  Engineering Systems Journal}, 2020.

\bibitem{mataric1994reward}
M.~J. Mataric, ``Reward functions for accelerated learning,'' in \emph{Machine
  learning proceedings}, 1994.

\bibitem{shahriari2017generic}
B.~Shahriari, ``Generic online learning for partial visible \& dynamic
  environment with delayed feedback,'' Ph.D. dissertation, San Jose State
  University, 2017.

\bibitem{feigenbaum2007distributed}
J.~Feigenbaum \emph{et~al.}, ``Distributed algorithmic mechanism design,'' in
  \emph{Algorithmic Game Theory}.\hskip 1em plus 0.5em minus 0.4em\relax
  Cambridge University Press, 2007.

\bibitem{pathak2017curiosity}
D.~Pathak \emph{et~al.}, ``Curiosity-driven exploration by self-supervised
  prediction,'' in \emph{ICML}, 2017.

\bibitem{biddinggame}
source, \url{https://github.com/DRACOsource/biddinggame}, 2021.

\bibitem{dulac2021challenges}
G.~Dulac-Arnold \emph{et~al.}, ``Challenges of real-world reinforcement
  learning: definitions, benchmarks and analysis,'' \emph{Machine Learning},
  pp. 1--50, 2021.

\bibitem{hester2013texplore}
T.~Hester and P.~Stone, ``Texplore: real-time sample-efficient reinforcement
  learning for robots,'' \emph{Machine learning}, 2013.

\bibitem{mann2018learning}
T.~A \emph{et~al.}, ``Learning from delayed outcomes with intermediate
  observations,'' \emph{arXiv preprint arXiv:1807.09387}, 2018.

\bibitem{hung2019optimizing}
C.-C. Hung \emph{et~al.}, ``Optimizing agent behavior over long time scales by
  transporting value,'' \emph{Nature communications}, 2019.

\bibitem{pathak18largescale}
Y.~Burda \emph{et~al.}, ``Large-scale study of curiosity-driven learning,'' in
  \emph{ICLR}, 2019.

\bibitem{chang2007no}
Y.-H. Chang, ``No regrets about no-regret,'' \emph{Artificial Intelligence},
  2007.

\bibitem{weinberg2004best}
M.~Weinberg \emph{et~al.}, ``Best-response multiagent learning in
  non-stationary environments,'' in \emph{AAMAS}, 2004.

\bibitem{bowling2002multiagent}
M.~Bowling \emph{et~al.}, ``Multiagent learning using a variable learning
  rate,'' \emph{Artificial Intelligence}, 2002.

\bibitem{heinrich2015fictitious}
J.~Heinrich \emph{et~al.}, ``Fictitious self-play in extensive-form games,'' in
  \emph{ICML}, 2015.

\bibitem{LESLIE2006285}
D.~S. Leslie \emph{et~al.}, ``Generalised weakened fictitious play,''
  \emph{Games and Economic Behavior}, 2006.

\bibitem{khaledi2016optimal}
M.~Khaledi \emph{et~al.}, ``Optimal bidding in repeated wireless spectrum
  auctions with budget constraints,'' in \emph{IEEE GLOBECOM}, 2016.

\bibitem{sutton2018reinforcement}
R.~S. Sutton and A.~G. Barto, \emph{Reinforcement learning: An
  introduction}.\hskip 1em plus 0.5em minus 0.4em\relax MIT press, 2018.

\bibitem{srivastava2015training}
R.~K. Srivastava \emph{et~al.}, ``Training very deep networks,'' in
  \emph{NeurIPS}, 2015.

\bibitem{yu2017seqgan}
L.~Yu \emph{et~al.}, ``Seqgan: Sequence generative adversarial nets with policy
  gradient,'' in \emph{AAAI}, 2017.

\bibitem{jain1984quantitative}
R.~K. Jain \emph{et~al.}, ``A quantitative measure of fairness and
  discrimination,'' \emph{Eastern Research Laboratory}, 1984.

\bibitem{rhoades1993herfindahl}
S.~A. Rhoades, ``The herfindahl-hirschman index,'' \emph{Fed. Res. Bull.},
  1993.

\end{thebibliography}

\end{document}